\documentclass[runningheads]{llncs}
\usepackage[utf8]{inputenc} 
\usepackage[T1]{fontenc}    
\usepackage{amsmath,amssymb,amsfonts}%
\usepackage{algorithm}%
\usepackage{algorithmicx}%
\usepackage{algpseudocode}%
\usepackage{multirow}
\usepackage{hyperref}       
\usepackage{cleveref}
\usepackage{url}            
\usepackage{booktabs}       
\usepackage{amsfonts}       
\usepackage{nicefrac}       
\usepackage{microtype}      
\usepackage{lipsum}		
\usepackage{graphicx}
\usepackage{xcolor}
\usepackage[numbers]{natbib}
\usepackage{doi}
\usepackage{rotating}

\definecolor{codegreen}{rgb}{0,0.6,0}
\definecolor{codegray}{rgb}{0.5,0.5,0.5}
\definecolor{codepurple}{rgb}{0.58,0,0.82}
\definecolor{backcolour}{rgb}{0.95,0.95,0.92}

\usepackage{listings}
\lstdefinestyle{mystyle}{
  backgroundcolor=\color{backcolour}, commentstyle=\color{codegreen},
  keywordstyle=\color{magenta},
  numberstyle=\tiny\color{codegray},
  stringstyle=\color{codepurple},
  basicstyle=\ttfamily\footnotesize,
  breakatwhitespace=false,         
  breaklines=true,                 
  captionpos=b,                    
  keepspaces=true,                 
  numbers=left,                    
  numbersep=5pt,                  
  showspaces=false,                
  showstringspaces=false,
  showtabs=false,                  
  tabsize=2
}
\begin{document}
\title{BSRBF-KAN: A combination of B-splines and Radial Basis Functions in Kolmogorov-Arnold Networks}
%
%
\author{Hoang-Thang Ta\inst{1}\orcidID{0000-0003-0321-5106}}
%
\authorrunning{Ta et al.}
%
\institute{Department of Information Technology, Dalat University, Vietnam
\\
\email{thangth@dlu.edu.vn}
%
}
\titlerunning{BSRBF-KAN}
\maketitle              
\begin{abstract}
In this paper, we propose BSRBF-KAN, a Kolmogorov Arnold Network (KAN) that combines B-splines and radial basis functions (RBFs) to fit input data during training. We perform experiments with BSRBF-KAN, multi-layer perception (MLP), and other popular KANs, including EfficientKAN, FastKAN, FasterKAN, and GottliebKAN over the MNIST and Fashion-MNIST datasets. BSRBF-KAN shows stability in 5 training runs with a competitive average accuracy of 97.55\% on MNIST and 89.33\% on Fashion-MNIST and obtains convergence better than other networks. We expect BSRBF-KAN to open up many combinations of mathematical functions for designing KANs. Our repo is publicly available at: \url {https://github.com/hoangthangta/BSRBF_KAN}.

\end{abstract}
\section{Introduction}


A recent work by \citet{liu2024kan} mentioned applying learnable activation functions as ``edges" to fit training data instead of using fixed ones as ``nodes," which are commonly used in multi-layer perceptrons (MLPs). The theory behind KANs relies on the Kolmogorov-Arnold representation theorem (KART), which states that a continuous function of multiple variables can be expressed as a combination of continuous functions of a single variable through additions \cite{kolmogorov1957representation}. KANs have been gaining more community attention in designing those networks and comparing their results with MLPs.


With the inspiration of KANs, many scientists have flocked to develop different types of KANs based on popular polynomials and basis functions. While these works focus on univariate functions to set up KANs, none have explored their combination. Therefore, we aim to combine B-splines~\cite{de1972calculating} and RBFs~\cite{buhmann2000radial} to build a combined KAN named BSRBF-KAN. The reason why we use them is their successful implementation in two popular KANs, EfficientKAN\footnote{https://github.com/Blealtan/efficient-kan} and FastKAN~\cite{li2024kolmogorov}. Additionally, they are continuous basis functions used for function approximation, providing features such as smoothness, adaptability, and local control. With the experiments on MNIST~\cite{deng2012mnist} and Fashion-MNIST~\cite{xiao2017fashion}, we show the stability and faster convergence of BSRBF-KAN compared to other networks. In this paper, our primary contribution is the development of a combined KAN, BSRBF-KAN, which we compare with other networks on two popular datasets (MNIST and Fashion-MNIST) to understand our KAN behaviors in the image classification problem.

The rest of this document is structured as follows: Section 2 contains related works on KART and KANs. Section 3 outlines our methodology, describing the KART and the design of the KAN architecture, as well as several currently popular KANs. Section 4 presents our experiments, where we compare BSRBF-KAN with MLP and other KANs using results from the MNIST and Fashion-MNIST datasets. This section also includes a misclassification analysis to understand which classes the models perform well on and an ablation analysis to identify the key components that contribute most to the performance of BSRBF-KAN. Next, Section 5 offers some limitations and how to handle them. Finally, Section 6 provides conclusions, comments, and our future work.

\section{Related Works}\label{sec_related_works}

In 1957, Kolmogorov provided proof to solve Hilbert’s 13th problem, demonstrating that a multivariate continuous function can be represented as a combination of single-variable functions and additions, known as KART~\cite{kolmogorov1957representation,braun2009constructive}. This theorem has been applied in numerous works to build neural networks~\cite{zhou2022treedrnet,sprecher2002space,koppen2002training,lin1993realization,leni2013kolmogorov,lai2021kolmogorov}. However, it has also sparked debates about KART's ability in designing neural networks. \citet{girosi1989representation} indicated that KART is irrelevant to neural networks because the inner function $\phi_{q,p}$ in \Cref{eq:kar} can be highly non-smooth~\cite{vitushkin1954hilbert}, preventing $f$ from being smooth, which is essential for generalization and noise stability in neural networks. In contrast, \citet{kuurkova1991kolmogorov} argued that KART is relevant to neural networks, demonstrating that linear combinations of affine functions can accurately approximate all one-variable functions with some sigmoidal functions. 

Despite the long history of application in neural networks, no works on KART had gained significant attention within the research community until the recent work by \citet{liu2024kan}. They recommended not adhering strictly to the KART but generalizing it to create KANs with more neurons and layers. As a result, KANs can surpass MLPs in both accuracy and interpretability on small-scale AI + Science tasks. However, the revival of KAN has also faced criticism by \citet{dhiman2024kan}, where KANs are described as MLPs with spline-based activation functions, unlike traditional MLPs, which have fixed activation functions. 

KANs are a fresh breeze that has blown into the scientific community since May 2024. Many KAN design works rely on popular mathematical functions, especially ones that can deal with curves such as B-Splines~\cite{de1972calculating} (Original KAN~\cite{liu2024kan}, EfficientKAN\footnote{https://github.com/Blealtan/efficient-kan}), Gaussian Radial Basis Functions or GRBFs (FastKAN~\cite{li2024kolmogorov}), Reflection SWitch Activation Function or RSWAF (FasterKAN~\cite{athanasios2024}),  Chebyshev polynomials (TorchKAN~\cite{torchkan}, Chebyshev KAN~\cite{ss2024chebyshev}), Legendre polynomials (TorchKAN~\cite{torchkan}), Fourier transform (FourierKAN\footnote{https://github.com/GistNoesis/FourierKAN/}, FourierKAN-GCF~\cite{xu2024fourierkan}) and other polynomial functions~\cite{teymoor2024exploring}.

Research in neural networks has focused on using linear combinations or activation functions to improve model performance. \citet{hashem1997optimal} extended optimal linear combinations (OLCs) by developing algorithms for selecting component networks, enhancing the generalization of OLCs. \citet{rodriguez2022replacing} explored alternative pooling functions to improve feature extraction in Convolutional Neural Networks. In another work, \citet{jie2021regularized} introduced principles for selecting activation functions, including combinations of ReLU and ELUs for improvements in time series forecasting and image compression. \citet{gutierrez2009combined} presented a hybrid neural network model that integrates various transfer projection functions with radial basis functions (RBF) in a feed-forward network's hidden layer, achieving better classification task performance.

\section{Methodology}
\label{sec:methodology}

\subsection{Kolmogorov-Arnold Representation Theorem}

A KAN relies on KART, which states that any multivariate continuous function $f$ defined on a bounded domain can be expressed using a finite number of continuous single-variable functions and additions~\cite{chernov2020gaussian,schmidt2021kolmogorov}. Given $\mathbf{x}={x_1,x_2,..,x_n}$ consisting of $n$ variables, a multivariate continuous function $f(\mathbf{x})$ is represented by:


\begin{equation}
\begin{aligned}
f(\mathbf{x}) = f(x_1, \ldots, x_n) = \sum_{q=1}^{2n+1} \Phi_q \left( \sum_{p=1}^{n} \phi_{q,p}(x_p) \right) 
\end{aligned}
\label{eq:kar}
\end{equation}
which contains two sums: outer sum and inner sum. The outer sum $\sum_{q=1}^{2n+1}$ will sum $2n+1$ of $\Phi_q$ ($\mathbb{R} \to \mathbb{R}$) terms. The inner one sums up $n$ terms for each $q$, and each term $\phi_{q,p}$ ($\phi_{q,p} \colon [0,1] \to \mathbb{R}$) represents a continuous function of a single variable $x_p$.

\subsection{The design of KANs}
\label{KAN_design}

Let's revisit an MLP characterized by affine transformations and non-linear functions.  From an input $\mathbf{x}$, the network performs the composition of weight matrices by layers (from layer $0$ to layer $L-1$) and the non-linearity $\sigma$ to generate the final output. 

\begin{equation}
\begin{aligned}
\text{MLP}(\mathbf{x}) &= (W_{L-1} \circ \sigma \circ W_{L-2} \circ \sigma \circ \cdots \circ W_1 \circ \sigma \circ W_0) \mathbf{x} \\
&= \sigma \left( W_{L-1} \sigma \left( W_{L-2} \sigma \left( \cdots \sigma \left( W_1 \sigma \left( W_0 \mathbf{x} \right) \right) \right) \right) \right)
\end{aligned}
\label{eq:mlp}
\end{equation}

\citet{liu2024kan} was inspired by KART to design KANs, but the authors advised generalizing it to more widths and depths. In \Cref{eq:kar}, we must search proper $\Phi_q$ and $\phi_{q,p}$ to solve the problem. A general KAN network consisting of $L$ layers takes $\mathbf{x}$ to generate the output as:

\begin{equation}
\begin{aligned}
\text{KAN}(\mathbf{x}) = (\Phi_{L-1} \circ \Phi_{L-2} \circ \cdots \circ \Phi_1 \circ \Phi_0)\mathbf{x}
\end{aligned}
\label{eq:kan}
\end{equation}
which $\Phi_{l}$ is the function matrix of the $l^{th}$ KAN layer or a set of pre-activations. Let denote the neuron $i^{th}$ of the layer $l^{th}$ and the neuron $j^{th}$ of the layer $l+1^{th}$. The activation function $\phi_{l,i,j}$ connects $(l, i)$ to $(l + 1, j)$:

\begin{equation}
\begin{aligned}
\phi_{l,j,i}, \quad l = 0, \cdots, L - 1, \quad i = 1, \cdots, n_l, \quad j = 1, \cdots, n_{l+1}
\end{aligned}
\label{eq:acti_funct}
\end{equation}
with $n_l$ is the number of nodes of the layer $l^{th}$. So now, the function matrix $\Phi_{l}$ can be represented as a matrix $n_{l+1} \times n_{l}$ of activations as:

\begin{equation}
\begin{aligned}
\mathbf{x}_{l+1} = 
\underbrace{\left(
\begin{array}{cccc}
\phi_{l,1,1}(\cdot) & \phi_{l,1,2}(\cdot) & \cdots & \phi_{l,1,n_l}(\cdot) \\
\phi_{l,2,1}(\cdot) & \phi_{l,2,2}(\cdot) & \cdots & \phi_{l,2,n_l}(\cdot) \\
\vdots & \vdots & \ddots & \vdots \\
\phi_{l,n_{l+1},1}(\cdot) & \phi_{l,n_{l+1},2}(\cdot) & \cdots & \phi_{l,n_{l+1},n_l}(\cdot)
\end{array}\right)}_{\Phi_{l}} \mathbf{x}_l
\label{eq:function_matrix}
\end{aligned}
\end{equation}

\subsection{Implementation of the current KANs}
\textbf{Original KAN} was implemented by \citet{liu2024kan} by using the residual activation function $\phi(x)$ as the sum of the base function and the spline function with their corresponding weight matrices $w_b$ and $w_s$:

\begin{equation}
\begin{aligned}
\phi(x) = w_b b(x) + w_s spline(x)
\end{aligned}
\label{eq:acti_funct_imp}
\end{equation}
where \(b(x)\) equals to $silu(x)$ and \(s(x)\) is expressed as a linear combination of B-splines. Each activation function is activated with \(w_s = 1\) and \(spline(x) \approx 0\), while \(w_b\) is initialized by using Xavier initialization.

\textbf{EfficientKAN} adopted the same approach as \citet{liu2024kan} but reworked the computation using B-spline basis functions followed by linear combination, reducing memory cost and simplifying computation\footnote{https://github.com/Blealtan/efficient-kan}. The authors replaced the incompatible L1 regularization on input samples with L1 regularization on weights. They also added learnable scales for activation functions and switched the base weight and spline scaler matrices to Kaiming uniform initialization, significantly improving performance on MNIST.

\textbf{FastKAN} can speed up training over EfficientKAN by using GRBFs to approximate the 3-order B-spline and employing layer normalization to keep inputs within the RBFs' domain~\cite{li2024kolmogorov}. These modifications simplify the implementation without sacrificing accuracy. The RBF has the formula:
\begin{equation}
\begin{aligned}
\phi(r) = e^{-\epsilon r^2}
\end{aligned}
\label{eq:gaussian_rbf}
\end{equation}
where $r = \|x - c\|$  is the Euclidean distance between a given point $x$ and a center point $c$, and $\epsilon$ ($epsilon > 0$) is a sharp parameter that controls the width of the Gaussian function. FastKAN uses a special form of RBFs, GRBFs where $\epsilon = \frac{1}{2}$ as~\cite{fornberg2011stable,li2024kolmogorov}:

\begin{equation}
\begin{aligned}
\phi_{\mathit{RBF}}(r) = \exp\left(-\frac{r^2}{2h^2}\right)
\end{aligned}
\label{eq:special_gaussian_rbf}
\end{equation}
and $h$ to control the width of the Gaussian function. Finally, the RBF network with $N$ centers can be shown as~\cite{li2024kolmogorov}:

\begin{equation}
\begin{aligned}
RBF(x) = \sum_{i=1}^{N} w_i \phi_{\mathit{RBF}}(r_i) = \sum_{i=1}^{N} w_i \exp\left(-\frac{||x - c_i||}{2h^2}\right)
\end{aligned}
\label{eq:rbf_network}
\end{equation}
where $w_i$ represents adjustable weights or coefficients, and $\phi$ denotes the radial basis function as in \Cref{eq:gaussian_rbf}.

\textbf{FasterKAN} outperforms FastKAN in both forward and backward processing speeds~\cite{athanasios2024}. It uses Reflectional Switch Activation Functions (RSWAFs), which are variants of RBFs and straightforward to compute due to their uniform grid structure. The RSWAF function $\phi(r)$ is shown as:

\begin{equation}
\begin{aligned}
\phi_{\mathit{RSWAF}}(r) = 1 - \left(\tanh\left(\frac{r}{h}\right)\right)^2
\end{aligned}
\label{eq:rswaf_funct}
\end{equation}

Then, the RSWAF network with $N$ centers will be:
\begin{equation}
\begin{aligned}
\mathit{RSWAF}(x) = \sum_{i=1}^{N} w_i \phi_{\mathit{RSWAF}}(r_i) = \sum_{i=1}^{N} w_i \left(1 - \left(\tanh\left(\frac{||x - c_i||}{h}\right)\right)^2\right)
\end{aligned}
\label{eq:rswaf_network}
\end{equation}

\textbf{GottliebKAN}, which is based on Gottlieb polynomials, showed the best result compared to the other 17 polynomial basis functions for setting up KANs on MNIST~\cite{teymoor2024exploring}. Gottlieb polynomials are a family of polynomials that arise in the study of the Bernoulli numbers and have applications in combinatorics and number theory~\cite{gottlieb1938concerning}. 
\citet{teymoor2024exploring} denoted Gottlieb polynomials as:
\begin{equation}
\begin{aligned}
G_n(x) = \sum_{k=0}^{n} \binom{n}{k} B_k(x) x^{n-k}
\end{aligned}
\label{eq:gottlieb}
\end{equation}
where $G_n(x)$ represents the $n$-th Gottlieb polynomial, $(n \; k)$ denotes the binomial coefficient "n choose k", $B_k(x)$ are the Bernstein basis polynomials. Finally, $x^{n-k}$ is the power term that adjusts the degree of $B_k(x)$ within the polynomial $G_n(x)$.

\subsection{BSRBF-KAN}
Inspired by successful implementations of previous KAN models (EfficientKAN and FastKAN), we developed BSRBF-KAN by combining B-splines and RBFs, which are widely recognized in computational mathematics for interpolation and approximation tasks. The RBFs we chose here are Gaussian ones because they are popular. Both B-splines and RBFs are smooth functions, ensuring continuity and derivatives up to a specified order, making them suitable for choosing proper $\Phi_q$ and $\phi_{q,p}$ in KANs. 

From an input $x$, the BSRBF function is represented as:
\begin{equation}
\begin{aligned}
\mathit{BSRBF}(x)  =  w_b b(x) + w_s (\mathit{BS}(x) + \mathit{RBF}(x)) 
\end{aligned}
\label{eq:bsrbf_funct}
\end{equation}
where $b(x)$ and $w_b$ are the base (linear) output and its base matrix. $\mathit{BS}(x)$ and $\mathit{RBF}(x)$ are B-Spline and RBF functions, and $w_s$ is the spline matrix. Additionally, we apply the BSRBF function to inputs normalized by layer normalization at each model layer. 

We also provide the Python code for the \texttt{forward()} function, in which a given input $\mathbf{x}$ is passed through layer normalization before returning the combination of its outputs. Given the goal of designing a combined KAN, our intuition suggests that the forward and backward speeds will not surpass those of individual KANs. 

\begin{lstlisting}[language=Python, caption=]
def forward(self, x):
        
        # layer normalization
        x = self.layernorm(x)
        # base
        base_output = F.linear(self.base_activation(x), self.base_weight)
        # b_splines
        bs_output = self.b_splines(x).view(x.size(0), -1)
        # rbfs
        rbf_output = self.rbf(x).view(x.size(0), -1)
        # combine
        bsrbf_output =  bs_output + rbf_output
        bsrbf_output = F.linear(bsrbf_output, self.spline_weight)
        return base_output + bsrbf_output
\end{lstlisting}

\section{Experiments}
\begin{figure*}[htbp]
  \centering

\includegraphics[scale=0.65]{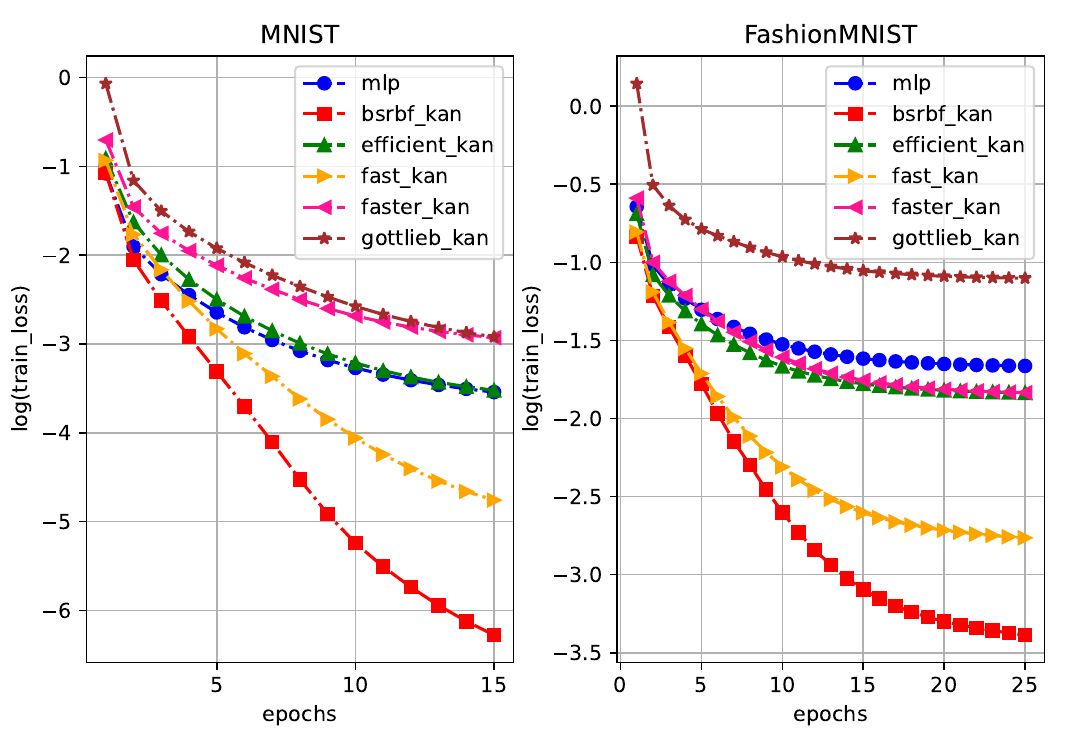}

  \centering
  \caption{The logarithmic values of training losses during a training run over 15 epochs on MNIST and 25 epochs on Fashion-MNIST.}
\label{fig:train_losses}
\end{figure*}

We conducted 5 independent training runs for each model on MNIST~\cite{deng2012mnist} and Fashion-MNIST~\cite{xiao2017fashion} to assess their overall performance more reliably. We chose these two datasets because they are popular, relatively simple, and small-scale, which allows us to test and perform experiments without using too many computer resources. By taking the best metric value from all runs, we aimed to mitigate the effects of variability in training and ensure that the evaluation reflects the models' peak capabilities. With the intention of designing simple networks, we only use activation functions, linear transformation, and layer normalization. 

All models have a network structure of (784, 64, 10), comprising 784 input neurons, 64 hidden neurons, and 10 output neurons corresponding to the 10 output classes (from 0 to 9). For GottliebKAN, we used a structure (784, 64, 64, 10) to make it the same as the original design\footnote{https://github.com/seydi1370/Basis\_Functions}. We trained the models with 15 epochs on MNIST and 25 epochs on Fashion-MNIST. Other hyperparameters are the same in all models, including \texttt{batch\_size=64}, \texttt{learning\_rate=1e-3}, \texttt{weight\_decay=1e-4}, \texttt{gamma=0.8}, \texttt{optimize=AdamW}, and \texttt{loss=CrossEntropy}. 

\Cref{fig:train_losses} demonstrates training losses of models represented by logarithm values in a certain training run. On both MNIST and Fashion-MNIST, the training loss values of BSRBF-KAN drop sharply continuously, which makes it the model with the smallest loss after the training process. While this accelerates the training, there is also a concern that the model might be overfitting. We examined both training and validation accuracies across datasets and found no anomalies. In contrast, GottliebKAN lost the least loss, which affected its performance, as shown in the next experiments.

\begin{table*}[ht]
	\caption{The best metric values in 5 training runs on MNIST and Fashion-MNIST.}
	\centering
	\begin{tabular}{p{2cm}p{2cm}p{1.3cm}p{1.3cm}p{1.3cm}p{1.5cm}p{1.4cm}}
            \hline
		\textbf{Dataset} &  \textbf{Model}   & \textbf{Train. Acc.} & \textbf{Val. Acc.} & \textbf{F1} & \textbf{Time (seconds)} & \textbf{\#Params} \\
    \hline 
    \multirow{6}{2cm}{\textbf{MNIST}}  &  BSRBF-KAN	&	\textbf{100.0}	& 97.63	& 97.6	& 222 & 459040  \\
&  FastKAN	& 	99.94	& 97.38 &	97.34	& 102 & 459114 \\
& FasterKAN	&	98.52 &	97.38 &	97.36 &	\textbf{93} & 408224 \\
& EfficientKAN	&	99.34 &	97.54 &	97.5 &	122 & 508160 \\
& GottliebKAN &	99.66 & \textbf{97.78} &	\textbf{97.74} 	& 269 & 219927 \\
& MLP & 99.42 &  97.69 &  97.66 & 273 &  \textbf{52512} \\
            \hline
            \multirow{6}{2cm}{\textbf{Fashion-MNIST}} & 	BSRBF-KAN	&	\textbf{99.3} &	89.59 &	89.54	& 219 & 459040  \\
 &  FastKAN	& 	98.27	& \textbf{89.62} &	\textbf{89.6}	& 160 & 459114 \\
&  FasterKAN	&	94.4 &	89.39 &	89.3 &	157 & 408224 \\
&  EfficientKAN	&	94.83 &	89.11 &	89.04 &	182 & 508160 \\
&  GottliebKAN &	93.79 &	87.69 &	87.61 	& 241 & 219927 \\
&  MLP & 93.58 &	88.51	& 88.44 & \textbf{147} &  \textbf{52512} \\
            \hline
            \multirow{6}{2.5cm}{\textbf{Average of MNIST + Fashion-MNIST}} & 	BSRBF-KAN	& \textbf{99.65}	& \textbf{93.61}	& \textbf{93.57}	& 220.5	& 459040
  \\
 &  FastKAN	& 	 99.11	& 93.50	& 93.47	& 131	& 459114 \\
&  FasterKAN	& 96.46	& 93.39	& 93.33	& \textbf{125}	& 408224
 \\
&  EfficientKAN& 97.09	& 93.33	& 93.27	& 152	& 508160
 \\
&  GottliebKAN & 96.73	& 92.74	& 92.68	& 255	& 219927
 \\
&  MLP & 96.50	& 93.10	& 93.05	& 210	& \textbf{52512}  \\
            \hline
             \multicolumn{6}{l}{Train. Acc = Training Accuracy, Val. Acc. = Validation Accuracy}  \\
             \multicolumn{6}{l}{\#Params = Parameters }  \\
             \hline
	\end{tabular}
	\label{tab:best_metric}
\end{table*}

\begin{table*}[ht]
	\caption{The average metric values in 5 training runs on MNIST and Fashion-MNIST.}
	\centering
	\begin{tabular}{p{1.7cm}p{2cm}p{2.2cm}p{2.2cm}p{2.2cm}p{1.2cm}}
            \hline
		\textbf{Dataset} &  \textbf{Model}   & \textbf{Train. Acc.} & \textbf{Val. Acc.} & \textbf{F1} & \textbf{Time (secs)} \\
    \hline 
	\multirow{6}{2cm}{\textbf{MNIST}} &	BSRBF-KAN	& \textbf{100.00 ± 0.00} &	97.55 ± 0.03 &	97.51 ± 0.03	& 231 \\
& FastKAN	&	99.94 ± 0.01 &	97.25 ± 0.03	& 97.21 ± 0.03 &	101 \\
& FasterKAN	&	98.48 ± 0.01 &	97.28 ± 0.06	& 97.25 ± 0.06 &	\textbf{93} \\
& EfficientKAN	&	99.37 ± 0.04 &	97.37 ± 0.07 &	97.33 ± 0.07	 &	120 \\
& GottliebKAN &	98.44 ± 0.61	& 97.19 ± 0.22	& 97.14 ± 0.23 	& 221 \\
& MLP & 99.44 ± 0.01 & \textbf{97.62 ± 0.03} & \textbf{97.59 ± 0.03} & 181 \\
            \hline
           \multirow{6}{2cm}{\textbf{Fashion-MNIST}} & BSRBF-KAN	& \textbf{99.19 ± 0.03}	& 89.33 ± 0.07	& 89.29 ± 0.07	& 211 \\
& FastKAN	&	98.19 ± 0.04	& \textbf{89.42 ± 0.07}	& \textbf{89.38 ± 0.07} &	162 \\
& FasterKAN	&	94.40 ± 0.01 &	89.26 ± 0.06 &	89.17 ± 0.07 &	154 \\
& EfficientKAN	&	94.76 ± 0.06 &	88.92 ± 0.08 &	88.85 ± 0.09 &	183 \\
& GottliebKAN &	90.66 ± 1.08	& 87.16 ± 0.24 &	87.07 ± 0.25 	& 238 \\
& MLP & 93.56 ± 0.05 &	88.39 ± 0.06 &	88.36 ± 0.05 & \textbf{148} \\
            \hline
           \multirow{6}{2.5cm}{\textbf{Average of MNIST + Fashion-MNIST}} & BSRBF-KAN 	& \textbf{99.60}	& \textbf{93.44}	& \textbf{93.40}	& 221 \\
& FastKAN 	& 99.07	& 93.34	& 93.30	& 131.5 \\
& FasterKAN 	& 96.44	& 93.27	& 93.21	& 123.5 \\
& EfficientKAN 	& 97.07	& 93.15	& 93.09	& 151.5 \\
& GottliebKAN 	& 94.55	& 92.18	& 92.11	& 229.5 \\
& MLP 	& 96.50	& 93.01	& 92.98	& 164.5 \\
\hline
             \multicolumn{5}{l}{Train. Acc = Training Accuracy, Val. Acc. = Validation Accuracy }  \\
             \hline
	\end{tabular}
	\label{tab:average_metric}
\end{table*}

\subsection{MNIST results analysis}
As shown in \Cref{tab:best_metric}, GottliebKAN is better than the other models, with a validation accuracy of 97.78\% and an F1 score of 97.74\%. We assume this because of its deeper network structure. MLP and BSRBF-KAN achieved second-best and third-best performances but took a long time to train. Conversely, FasterKAN had the shortest training time, but its performance slightly surpassed FastKAN. BSRBF-KAN is the only model with a training accuracy of 100\% after 15 epochs. If counting the results based on average values, MLP is the best model, as displayed in \Cref{tab:average_metric}, followed competitively by BSRBF-KAN. Despite achieving the highest accuracy values, GottliebKAN is unstable. 

\subsection{Fashion-MNIST results analysis}
The ease of achieving high accuracy of models trained on MNIST may not accurately reflect their actual performance. Thus, we experimented with the model training on Fashion-MNIST. As shown in \Cref{tab:best_metric} and \Cref{tab:average_metric}, FastKAN obtained the highest validation accuracy. At the same time, BSRBF-KAN followed competitively with and still kept the best convergence ability. MLP is the second-worst model, only performing better than GottliebKAN, but it has the fastest training time and the fewest parameters. However, when considering training time, we see that FasterKAN's training time is only slightly longer than MLP's, while achieving nearly 0.9\% higher validation accuracy, according to the results in \Cref{tab:best_metric}. 

\subsection{MNIST + Fashion-MNIST results analysis}
When calculating the average values on MNIST and Fashion-MNIST, we clearly see that BSRBF-KAN achieved the best performance and also the best convergence, according to \Cref{tab:best_metric} and \Cref{tab:average_metric}. Furthermore, all KANs except for GottliebKAN surpassed MLP. This indicates the positive signal of using KANs to obtain better accuracy values in model training, at least in these experiments.

\subsection{Misclassification Analysis}
\begin{figure*}[htbp]
  \centering

\includegraphics[scale=0.48]{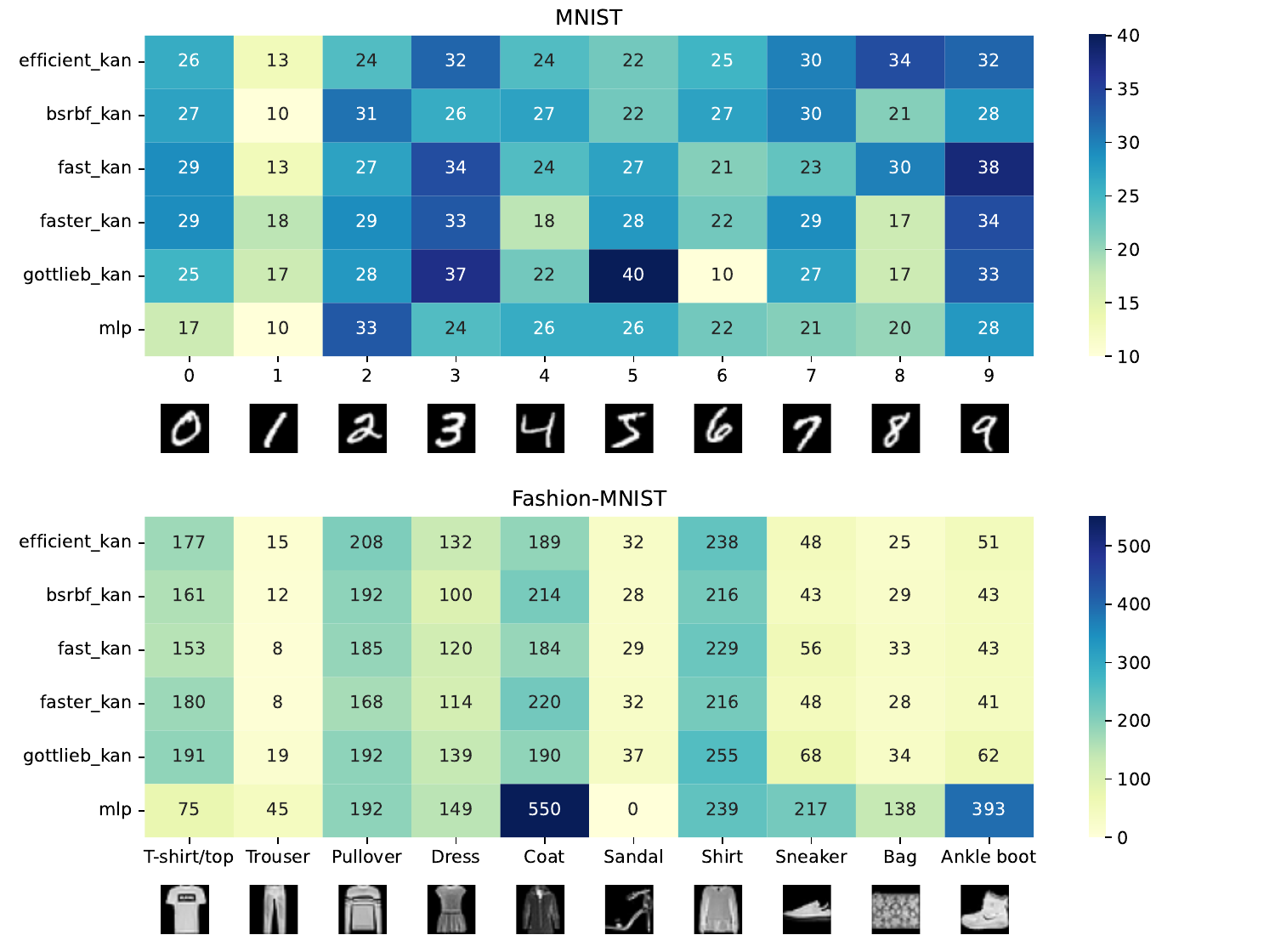}

  \centering
  \caption{Heatmap of the misclassified images in the test set by models over MNIST and Fashion-MNIST.}
\label{fig:misclass}
\end{figure*}

In this section, we experimented with the misclassifications in the MNIST and Fashion-MNIST datasets, as illustrated in the heatmaps in \Cref{fig:misclass}. Each model indicates the number of misclassified images by class in the test sets. This experiment aims to elucidate the specific data classes where each model excels, generating an understanding that will inform subsequent refinements in model design.

For MNIST, classes 2, 3, and 9 are generally challenging for most models, while class 1 is the easiest, exhibiting the fewest misclassifications. The BSRFB-KAN model shows a relatively small gap compared to other models; however, it does have higher misclassification rates in certain classes, specifically classes 0, 2, 4, 6, and 7. Both FasterKAN and GottliebKAN demonstrate strong performance in classes 1 and 8, with GottliebKAN also performing well in class 6.

In Fashion-MNIST, while all KANs exhibit similar misclassification rates across classes, the MLP model struggles particularly with the "Coat" and "Ankle boot" classes. Classes "Pullover," "Coat," and "Shirt" present classification challenges for all models due to their visual similarities. Notably, the MLP model makes no errors in classifying images from the "Sandal" class and has the fewest errors in classifying "T-shirt/top." Based on these analyses, we suggest that combining MLP and KANs, while addressing their flaws, may help in designing new models to improve performance.

\subsection{Ablation Study}
Several ablation experiments were performed to help us understand the impact of various components within BSRBF-KAN on overall model performance. \Cref{tab:ablation_study} presents the metric values for different combinations of BSRBF-KAN components. Initially, we trained the model with all components included, followed by training variants with some components removed. The results indicate that the base output and layer normalization are key for the model's performance, as their both absence leads to the worst results.

While B-splines and RBFs can enhance performance, removing RBFs has less impact on the model than B-splines. Interestingly, when both are absent, the model becomes an MLP, which exhibits slightly better validation accuracy and F1 scores than the complete BSRBF-KAN model on MNIST, but on Fashion-MNIST, this is pretty bad. Thus, we believe that retaining B-splines and RBFs remains beneficial for the model's performance.

\begin{table*}[ht]
	\caption{The performance of BSRBF-KAN by different components on MNIST and Fashion-MNIST.}
	\centering
	\begin{tabular}{p{2cm}p{4cm}p{2cm}p{1.5cm}p{1.5cm}}
            \hline
		\textbf{Dataset} &  \textbf{Components}   & \textbf{Train. Acc.} & \textbf{Val. Acc.} & \textbf{F1} \\
    \hline 
	\multirow{6}{2cm}{\textbf{MNIST}} &	Full	& 100.0	& 97.53	& 97.49  \\
          &  No BS	&  - 0.01 & - 0.21	& - 0.21  \\
          &  No RBF	&  - 0.003   & - 0.14	& - 0.14  \\
          &  No BS + No RBF = MLP &  - 0.55    & + 0.05	& + 0.06  \\
          &  No BO	& - 0.01	& - 0.41	& - 0.41  \\
          & No LN	&  - 1.7 & - 1.5 & - 1.52  \\
          &  No BO + No LN &  \textbf{- 5.16} & \textbf{- 5.97} & \textbf{- 6.08}  \\
            \hline
          \multirow{6}{2cm}{\textbf{Fashion-MNIST}} &  Full & 99.39 & 89.43 & 89.4 \\
          &  No BS & - 0.16 & - 0.39 & - 0.37 \\
          &  No RBF & - 0.34 & - 0.14 & - 0.15 \\
          &  No BS + No RBF = MLP& -5.86 & -0.81 & -0.78 \\
          &  No BO & - 0.04 & - 0.26 & - 0.29 \\
          &  No LN	& - 6.12 & - 1.1 & - 1.13 \\
          &  No BO + No LN &  \textbf{-6.23} & \textbf{-3.29} & \textbf{-3.38} \\

            \hline
             \multicolumn{4}{l}{Train. Acc = Training Accuracy, Val. Acc. = Validation Accuracy}  \\
             \multicolumn{4}{l}{No BS = No B-spline, No RBF = No Radial Basis Function}  \\
             \multicolumn{4}{l}{No BO = No Base Output, No LN = No Layer Normalization}  \\
             \hline
	\end{tabular}
	\label{tab:ablation_study}
\end{table*}

\section{Limitation}
The first limitation of this study is that experiments were conducted solely on relatively simple datasets, specifically MNIST and Fashion-MNIST, leaving the effectiveness of BSRBF-KAN on larger, more complex datasets unexplored. Secondly, performance comparisons among KANs, including BSRBF-KAN against MLPs, may be impacted by model architecture (e.g., 784, 64, 10) rather than an equivalent parameter count, potentially affecting evaluation fairness. Lastly, combining B-splines and Radial Basis Functions was chosen based on observed performance in prior KANs (e.g., EfficientKAN, FastKAN) without a formalized mathematical justification or deep analyses. Addressing these limitations would require testing a broader range of datasets and tasks, designing models with equivalent parameter counts for balanced comparisons, and exploring the theoretical basis for function combinations through mathematical and qualitative analyses.

\section{Conclusion}
We introduced BSRBF-KAN, a novel KAN that integrates B-splines and RBFs to fit input data in the training process. We did not intend to make BSRBF-KAN better than MLP and other KANs by optimizing hyperparameters; instead, we built it by default to understand aspects of KANs combination in training models. In the experiments, we trained BSRBF-KAN, MLP, and other KANs on the MNIST and Fashion-MNIST datasets. BSRBF-KAN demonstrated competitive performance and rapid convergence compared to other models. Furthermore, it is the best model if counting on the average values of both datasets. However, its high convergence rate can lead to overfitting, which can be mitigated by dropout and other techniques. Furthermore, we conducted a misclassification analysis to examine issues with misclassified images in BSRBF-KAN, providing some information for designing better future models, as well as an ablation study to identify the components most critical (e.g., layer normalization and base output) to its performance.

All models are generally stable, except for GottliebKAN. Although GottliebKAN achieved the highest validation accuracy on MNIST, it performed the worst on Fashion-MNIST. Similarly, while MLP achieved the highest average validation accuracy on MNIST, it underperformed compared to other KANs on Fashion-MNIST. Therefore, it is crucial to experiment with our approach on other datasets in the future to gain a more comprehensive understanding. We will also continue to explore the combination of KANs and their components when designing new KAN architectures to enhance model performance across a wider range of problems. 


%
%
%
%
%

\bibliographystyle{splncs04nat}
\bibliography{references}

\begin{thebibliography}{31}
\providecommand{\natexlab}[1]{#1}
\providecommand{\url}[1]{\texttt{#1}}
\providecommand{\urlprefix}{URL }
\expandafter\ifx\csname urlstyle\endcsname\relax
  \providecommand{\doi}[1]{doi:\discretionary{}{}{}#1}\else
  \providecommand{\doi}{doi:\discretionary{}{}{}\begingroup \urlstyle{rm}\Url}\fi

\bibitem[{Bhattacharjee(2024)}]{torchkan}
Bhattacharjee, S.S.: Torchkan: Simplified kan model with variations. \url{https://github.com/1ssb/torchkan/} (2024)

\bibitem[{Braun and Griebel(2009)}]{braun2009constructive}
Braun, J., Griebel, M.: On a constructive proof of kolmogorov’s superposition theorem. Constructive approximation \textbf{30}, 653--675 (2009)

\bibitem[{Buhmann(2000)}]{buhmann2000radial}
Buhmann, M.D.: Radial basis functions. Acta numerica \textbf{9}, 1--38 (2000)

\bibitem[{Chernov(2020)}]{chernov2020gaussian}
Chernov, A.V.: Gaussian functions combined with kolmogorov’s theorem as applied to approximation of functions of several variables. Computational Mathematics and Mathematical Physics \textbf{60}, 766--782 (2020)

\bibitem[{De~Boor(1972)}]{de1972calculating}
De~Boor, C.: On calculating with b-splines. Journal of Approximation theory \textbf{6}(1), 50--62 (1972)

\bibitem[{Delis(2024)}]{athanasios2024}
Delis, A.: Fasterkan. \url{https://github.com/AthanasiosDelis/faster-kan/} (2024)

\bibitem[{Deng(2012)}]{deng2012mnist}
Deng, L.: The mnist database of handwritten digit images for machine learning research [best of the web]. IEEE signal processing magazine \textbf{29}(6), 141--142 (2012)

\bibitem[{Dhiman(2024)}]{dhiman2024kan}
Dhiman, V.: Kan: Kolmogorov--arnold networks: A review. \url{https://vikasdhiman.info/reviews/KAN_a_review.pdf} (2024)

\bibitem[{Fornberg et~al.(2011)Fornberg, Larsson, and Flyer}]{fornberg2011stable}
Fornberg, B., Larsson, E., Flyer, N.: Stable computations with gaussian radial basis functions. SIAM Journal on Scientific Computing \textbf{33}(2), 869--892 (2011)

\bibitem[{Girosi and Poggio(1989)}]{girosi1989representation}
Girosi, F., Poggio, T.: Representation properties of networks: Kolmogorov's theorem is irrelevant. Neural Computation \textbf{1}(4), 465--469 (1989)

\bibitem[{Gottlieb(1938)}]{gottlieb1938concerning}
Gottlieb, M.J.: Concerning some polynomials orthogonal on a finite or enumerable set of points. American Journal of Mathematics \textbf{60}(2), 453--458 (1938)

\bibitem[{Guti{\'e}rrez et~al.(2009)Guti{\'e}rrez, Herv{\'a}s, Carbonero, and Fern{\'a}ndez}]{gutierrez2009combined}
Guti{\'e}rrez, P.A., Herv{\'a}s, C., Carbonero, M., Fern{\'a}ndez, J.C.: Combined projection and kernel basis functions for classification in evolutionary neural networks. Neurocomputing \textbf{72}(13-15), 2731--2742 (2009)

\bibitem[{Hashem(1997)}]{hashem1997optimal}
Hashem, S.: Optimal linear combinations of neural networks. Neural networks \textbf{10}(4), 599--614 (1997)

\bibitem[{Jie et~al.(2021)Jie, Gao, Vasnev, and Tran}]{jie2021regularized}
Jie, R., Gao, J., Vasnev, A., Tran, M.n.: Regularized flexible activation function combination for deep neural networks. In: 2020 25th International Conference on Pattern Recognition (ICPR), pp. 2001--2008, IEEE (2021)

\bibitem[{Kolmogorov(1957)}]{kolmogorov1957representation}
Kolmogorov, A.N.: On the representation of continuous functions of many variables by superposition of continuous functions of one variable and addition. In: Doklady Akademii Nauk, vol. 114, pp. 953--956, Russian Academy of Sciences (1957)

\bibitem[{K{\"o}ppen(2002)}]{koppen2002training}
K{\"o}ppen, M.: On the training of a kolmogorov network. In: Artificial Neural Networks—ICANN 2002: International Conference Madrid, Spain, August 28--30, 2002 Proceedings 12, pp. 474--479, Springer (2002)

\bibitem[{Kůrková(1991)}]{kuurkova1991kolmogorov}
Kůrková, V.: Kolmogorov's theorem is relevant. Neural computation \textbf{3}(4), 617--622 (1991)

\bibitem[{Lai and Shen(2021)}]{lai2021kolmogorov}
Lai, M.J., Shen, Z.: The kolmogorov superposition theorem can break the curse of dimensionality when approximating high dimensional functions. arXiv preprint arXiv:2112.09963  (2021)

\bibitem[{Leni et~al.(2013)Leni, Fougerolle, and Truchetet}]{leni2013kolmogorov}
Leni, P.E., Fougerolle, Y.D., Truchetet, F.: The kolmogorov spline network for image processing. In: Image Processing: Concepts, Methodologies, Tools, and Applications, pp. 54--78, IGI Global (2013)

\bibitem[{Li(2024)}]{li2024kolmogorov}
Li, Z.: Kolmogorov-arnold networks are radial basis function networks. arXiv preprint arXiv:2405.06721  (2024)

\bibitem[{Lin and Unbehauen(1993)}]{lin1993realization}
Lin, J.N., Unbehauen, R.: On the realization of a kolmogorov network. Neural Computation \textbf{5}(1), 18--20 (1993)

\bibitem[{Liu et~al.(2024)Liu, Wang, Vaidya, Ruehle, Halverson, Solja{\v{c}}i{\'c}, Hou, and Tegmark}]{liu2024kan}
Liu, Z., Wang, Y., Vaidya, S., Ruehle, F., Halverson, J., Solja{\v{c}}i{\'c}, M., Hou, T.Y., Tegmark, M.: Kan: Kolmogorov-arnold networks. arXiv preprint arXiv:2404.19756  (2024)

\bibitem[{Rodriguez-Martinez et~al.(2022)Rodriguez-Martinez, Lafuente, Santiago, Dimuro, Herrera, and Bustince}]{rodriguez2022replacing}
Rodriguez-Martinez, I., Lafuente, J., Santiago, R.H., Dimuro, G.P., Herrera, F., Bustince, H.: Replacing pooling functions in convolutional neural networks by linear combinations of increasing functions. Neural Networks \textbf{152}, 380--393 (2022)

\bibitem[{Schmidt-Hieber(2021)}]{schmidt2021kolmogorov}
Schmidt-Hieber, J.: The kolmogorov--arnold representation theorem revisited. Neural networks \textbf{137}, 119--126 (2021)

\bibitem[{Sprecher and Draghici(2002)}]{sprecher2002space}
Sprecher, D.A., Draghici, S.: Space-filling curves and kolmogorov superposition-based neural networks. Neural Networks \textbf{15}(1), 57--67 (2002)

\bibitem[{SS(2024)}]{ss2024chebyshev}
SS, S.: Chebyshev polynomial-based kolmogorov-arnold networks: An efficient architecture for nonlinear function approximation. arXiv preprint arXiv:2405.07200  (2024)

\bibitem[{Teymoor~Seydi(2024)}]{teymoor2024exploring}
Teymoor~Seydi, S.: Exploring the potential of polynomial basis functions in kolmogorov-arnold networks: A comparative study of different groups of polynomials. arXiv e-prints pp. arXiv--2406 (2024)

\bibitem[{Vitushkin(1954)}]{vitushkin1954hilbert}
Vitushkin, A.: On hilbert’s thirteenth problem. In: Dokl. Akad. Nauk SSSR, vol.~95, pp. 701--704 (1954)

\bibitem[{Xiao et~al.(2017)Xiao, Rasul, and Vollgraf}]{xiao2017fashion}
Xiao, H., Rasul, K., Vollgraf, R.: Fashion-mnist: a novel image dataset for benchmarking machine learning algorithms. arXiv preprint arXiv:1708.07747  (2017)

\bibitem[{Xu et~al.(2024)Xu, Chen, Li, Yang, Wang, Hu, and Ngai}]{xu2024fourierkan}
Xu, J., Chen, Z., Li, J., Yang, S., Wang, W., Hu, X., Ngai, E.C.H.: Fourierkan-gcf: Fourier kolmogorov-arnold network--an effective and efficient feature transformation for graph collaborative filtering. arXiv preprint arXiv:2406.01034  (2024)

\bibitem[{Zhou et~al.(2022)Zhou, Zhu, Wang, Ma, Wen, Sun, and Jin}]{zhou2022treedrnet}
Zhou, T., Zhu, J., Wang, X., Ma, Z., Wen, Q., Sun, L., Jin, R.: Treedrnet: a robust deep model for long term time series forecasting. arXiv preprint arXiv:2206.12106  (2022)

\end{thebibliography}
\end{document}